\pdfoutput=1

\documentclass[11pt]{article}
\usepackage{fontawesome}
\usepackage{titlesec}

\usepackage{acl}
\usepackage{times}
\usepackage{latexsym}
\usepackage{graphicx}
\usepackage{booktabs}
\usepackage{placeins}
\usepackage{amsmath}

\usepackage[T1]{fontenc}

\usepackage[utf8]{inputenc}

\usepackage{microtype}

\usepackage{inconsolata}

\title{Deep Learning-based Computational Job Market Analysis:\\ A Survey on Skill Extraction and Classification from Job Postings}

\author{Elena Senger$^{1,3}$, Mike Zhang$^2$, Rob van der Goot$^2$, Barbara Plank$^{1,2}$\\
       $^1$MaiNLP, Center for Information and Language Processing, LMU Munich, Germany\\
       $^2$Department of Computer Science, IT University of Copenhagen, Denmark\\ 
       $^3$Fraunhofer Center for International Management and Knowledge Economy IMW, Germany\\
       \texttt{elena.senger@cis.lmu.de},  \texttt{\{mikz, robv\}@itu.dk}, \texttt{b.plank@lmu.de}
       }

\begin{document}

\maketitle
\begin{abstract} 
Recent years have brought significant advances to Natural Language Processing (NLP), which enabled fast progress in the field of \emph{computational job market analysis}. Core tasks in this application domain are \emph{skill extraction and classification} from job postings. Because of its quick growth and its interdisciplinary nature, there is no exhaustive assessment of this emerging field. This survey aims to fill this gap by providing a comprehensive overview of deep learning methodologies, datasets, and terminologies specific to NLP-driven skill extraction and classification. Our comprehensive cataloging of publicly available datasets addresses the lack of consolidated information on dataset creation and characteristics. 
Finally, the focus on terminology addresses the current lack of consistent definitions for important concepts, such as hard and soft skills, and terms relating to skill extraction and classification.

\end{abstract}

\section{Introduction}
\label{ch:Introduction}

Skill extraction and classification has recently been the subject of an increased amount of
interest \citep{zhang-etal-2023-escoxlm, clavié2023large}, which shows in a high number of publications, driven by the advances in natural language processing (NLP) technology. For instance, through large language models (LLMs) the low resource tasks of skill extraction can be approached by using synthetic training data \citep{clavié2023large, decorte2023extreme}.
Surveys regarding skill extraction are emerging \citep{Khaouja.2021, Papoutsoglou.2019}, nevertheless, a comprehensive overview from an NLP perspective is still lacking---a gap we aim to fill in this survey.
Our contributions are:
\begin{itemize}
    \itemsep0em
    \item Firstly, we aim to address the lack of standardized terminology in the field, bringing clarity to terms like hard and soft skills, as well as phrases related to skill extraction and classification.
    \item Additionally, this survey is the first to examine various publicly accessible datasets and sheds light on their creation methodologies.
    \item In contrast to prior surveys, we adopt an NLP-centric focus, with a deep dive into the latest advancements of neural methods for skill extraction and classification. 
\end{itemize}

While prior surveys exists, they focus typically on \emph{Skill count} and \emph{Topic modeling} methods for extracting skills. Skill count is performed manually or by matching n-grams with a skill base.  Topic modeling is an unsupervised method utilizing word distributions to identify underlying topics in documents. Due to primary statistical basis and lack of defined skill spans or labels, topic modeling, as well as skill count, methods are not covered in this survey. For further details on skill count, see \citet{Khaouja.2021} and \citet{Ternikov.2022}, and for topic modeling, please refer to \citet{Khaouja.2021}, \citet{Ternikov.2022} and \citet{Ao2023}.


\paragraph{Research Methodology} For our search strategy we used several academic databases including the ACL Anthology, Google Scholar, arXiv, IEEE, ACM, Science Direct, and Springer Link. The primary search terms were ``skill extraction'' and ``job''. To refine the search, we added terms like ``deep learning'', ``machine learning'', or ``natural language processing'' to our query for Google Scholar and Science Direct databases. This yielded the inclusion of 26 publications on neural skill extraction from job postings (JPs) that were published before November 2023. 

\section{Other Surveys} 
Previous surveys provide a foundation for our survey. Notable contributions include works from the social sciences, in particular, by~\citet{Napierala2023} in the ``Handbook of Computational Social Science for Policy'' (Chapter 13). It focuses on changing skills in a dynamic world from a social science perspective. Moreover, \citet{Papoutsoglou.2019} focus on studies regarding the software engineering labor market. Besides JPs, they research other sources like social networks or Q\&A sites. Lastly, the survey by \citet{Khaouja.2021} on skill identification from JPs is the closest to this survey. It overviews papers using methodologies such as skill counts, topic modeling, skill embeddings, and other machine learning-based methods. With this survey, we steer away from manual and topic modeling approaches to delve deeply into recent extraction methodologies and deep learning-based innovations.

\section{Skill-related Terminology} \label{chap:Terminology}

The terms \textit{skill extraction}, \textit{identification} \citep{li2023skillgpt}, \textit{detection} \citep{beauchemin2022fijo}, \textit{standardization} \citep{li2023skillgpt} and \textit{classification} are used differently, sometimes interchangeably, and describe the same or different tasks. 
We provide the following definition (See an example in Table \ref{tab:terminology} in the Appendix):
\begin{itemize}
    \itemsep0em
    \item \textbf{Skill Extraction (\(E\))}:  as a generic (parent) category for retrieving skill-related information. Skill extraction \(E: \text{JP} \rightarrow (S) \), where \(E\) maps a job posting (JP) to a set of skills \(S\). 
    \item \textbf{Skill Identification/Detection} (\(I\)): as the process of extracting skills without any pre-defined labels. It can be represented as \(I: \text{JP} \rightarrow S\), where skills, especially skill spans, are extracted from JPs. It can also be formalized as a classification problem,  \(I: \text{Span} \rightarrow \{0, 1\}\), to determine whether a given span in a JP represents a skill (1) or not (0).
    \item \textbf{Skill Extraction with Coarse Labels} (\(E_C\)): as identifying broader categories of skill spans. It is formalized as \(E_C: \text{JP} \rightarrow \{SC_1, SC_2, \dots, SC_n\}\), where each \(SC_i\) represents a skill span with a coarse label. 
    \item \textbf{Skill Standardization} (\(Std\)): as the normalization process of skill terms, formalized as \(Std: S \rightarrow S'\), mapping an initial set of skills \(S\) to a standardized set \(S'\). 
    \item \textbf{Direct Skill Classification} (\(C_D\)):  as mapping skills to a predefined skill base for assigning fine-grained labels. This process can be formalized as \(C_D: S \rightarrow L\), where \(C_D\) maps a set of already extracted skills \(S\)
    to a set of fine-grained labels \(L\).
    \item \textbf{Skill Classification with Extraction} (\(C_E\)):  as mapping JPs to a predefined skill base for assigning fine-grained labels. This process can be formalized as \(C_E: JP \rightarrow L\), where \(C_E\) maps a set of already extracted skills \(S\) entire JP or raw JP snippets
    to a set of fine-grained labels \(L\). 
    
\end{itemize}


Given these definitions, the skill extraction step can happen at different levels of \textbf{granularity} (of the input). Some works extract skills per JP (\(E_{JP}\), the overall document), per sentence (\(E_{sentence}\)) or per n-gram (\(E_{n-gram}\)). A skill span (\(E_{span}\))is a continuous n-gram sequence that capture a skill. 

A \textbf{skill base} (\(B\)) is a knowledge base containing skill entities and terminology. A taxonomy is a hierarchically structured skill base, while ontologies provide a structure via relationships between concepts \citep{Khaouja.2021}. Several works use the term ``skill dictionary'' for a skill base, most often referring to an unstructured skill base or a list of skills~\citep{Gugnani_Misra_2020, Yao.2022}. Two popular publicly-available skill bases, created by domain experts, and are frequently used and maintained are the European Skills, Competences, Qualifications and Occupations (ESCO;~\citealp{Vrang.2014}) taxonomy and the US Occupational Information Network (O*NET;~\citealp{national2010database}). We refer to \citet{Khaouja.2021} for more examples of skill bases.

\section{What are Skills? On Skill Definitions}\label{chap:definitions}

Understanding the concept of a \emph{skill} is pivotal in the field of skill extraction. In this section, we investigate several definitions of skills by various publications and institutions, aiming to identify commonalities and distinctions across different sources, which is crucial for establishing a common ground in this emerging field.

The concept of \emph{skill} can be seen as one broad concept \citep{green-etal-2022-development, Wild.2021, Fang.2023} or split into subclasses, with multiple possibilities for the split. In the latest version of the ESCO taxonomy the ``skill pillar'' is divided into four categories: ``Transversal skills'', ``Skills'',`` Knowledge'' and ``Language skills and knowledge''.\footnote{\url{https://esco.ec.europa.eu/en/classification/skill_main}} O*NET is structured in six domains  \citep{national2010database}, the domain most fitting for skill extraction from JP is ``Worker Requirements''. This domain entails four subcategories: basic skills, cross-functional skills, knowledge, and education.\footnote{\label{note_ONET} \url{https://www.onetcenter.org/content.html}} But publications considered in this survey that define skills, mainly distinguish between hard and soft skills \citep{Tamburri.2020, beauchemin2022fijo, Sayfullina.2018}, which is therefore also the separation used in this survey.

\paragraph{Hard Skills}

\citet{Tamburri.2020} delineate hard skills as professional competencies, activities, or knowledge pertinent to organizational functions, processes, and roles, essential for the successful completion of specific tasks. This definition emphasizes the practicality and functionality of hard skills within a professional setting. Aligning with this, the study by \citet{beauchemin2022fijo} views hard skills as task-oriented technical competencies, drawing upon \citet{Lyu.2021} to define them as formal technical abilities for performing certain tasks. Furthermore, \citet{Gugnani_Misra_2020} expand on this perspective by incorporating technological terminologies for skill identification and therefore integrating knowledge as a fundamental component of hard skills.

By incorporating knowledge as a component of hard skills, the definitions of hard skills and knowledge categories of O*NET and ESCO can be combined. O*NET's definition of hard skills states that they are developed abilities that enable learning or knowledge acquisition, coupled with their definition of knowledge as ``Organized sets of principles and facts applying in general domains''.\footnote{See footnote 2.} This comprehensive definition underscores not only technical proficiency but also the ability to adapt and apply knowledge. Similarly, ESCO, referencing the European Qualifications Framework, defines skills as ``the ability to apply knowledge and use know-how to complete tasks and solve problems'', while defining knowledge as ``the outcome of the assimilation of information through learning''.\footnote{\url{https://esco.ec.europa.eu/en/about-esco/escopedia/escopedia/knowledge} and \url{https://esco.ec.europa.eu/en/about-esco/escopedia/escopedia/skill}}

In conclusion, we define hard skills as a wide variety of professional abilities, ranging from measurable technical skills to the more general capacity for learning and effectively applying knowledge. They are quantifiable and teachable competencies, predominantly technical, yet intrinsically linked to the ability to adapt and apply them in diverse professional scenarios.

\paragraph{Soft Skills}

\citet{Sayfullina.2018}, referencing the Collins dictionary \citep{collins_dictionary}, views soft skills as innate, non-technical qualities highly sought after in employment, diverging from reliance on acquired knowledge.
In a more social context, \citet{Tamburri.2020} characterizes soft skills as encompassing personal, emotional, social, or intellectual aspects, further known as behavioral skills or competencies. Echoing this sentiment, \citet{beauchemin2022fijo}, drawing from \citet{Lyu.2021}, identifies soft skills as a variety of personal attributes and behaviors crucial for effective workplace interaction, collaboration, and adaptability.

Adding to these perspectives, ESCO characterizes soft skills as \emph{transversal skills}, highlighting their wide applicability across various occupations and sectors and their fundamental role in individual growth.\footnote{\url{https://esco.ec.europa.eu/en/about-esco/escopedia/escopedia/transversal-knowledge-skills-and-competences}} Similarly, O*NET classifies these skills under Cross-Functional Skills, defining them as developed capacities that enhance the performance of activities common across different jobs, encompassing areas like Social Skills and Complex Problem Solving Skills.\footnote{See footnote 2.}
Both sources underscore the universal relevance of soft skills.

These previous definitions lead to our converged definition that soft skills cover a vast array of personal, social, and intellectual competencies, all of which are indispensable for successful interpersonal engagement and personal development in professional settings.

\section{Operationalization of Skill Definitions}
In this section, we explore various methodologies for operationalizing skill definitions in skill extraction and classification research.
\paragraph{Using a Skill Base}
By using a given skill base, a pre-defined definition of the concept of skills is provided by the authors of the skill base. Numerous studies employ established skill bases such as the ESCO taxonomy \citep{zhang-etal-2023-escoxlm, zhang-etal-2022-kompetencer, clavié2023large, decorte2023extreme, Decorte.2022} or O*NET \citep{Gugnani_Misra_2020}. However, it is often ambiguous whether these studies use all or only specific subcategories \citep{li2023skillgpt, Decorte.2022, Gugnani_Misra_2020}. Some papers mention explicitly the use of all subclasses \citep{zhang-etal-2022-kompetencer,zhang-etal-2022-skillspan,gnehm-etal-2022-fine} other times it can be inferred from the number of skill spans used \citep{clavié2023large,decorte2023extreme}. However, one should note that the interpretations of ESCO definitions differ based on the ESCO version and authors' perspective. \citet{zhang-etal-2022-skillspan, zhang-etal-2022-kompetencer} used ESCO version 1.0 with a different soft skill category than discussed in Section \ref{chap:definitions} and implemented two labels: ``knowledge'' aligns with ESCO's ``Knowledge'' category, and ``Skills'' as a fusion of the hard and soft skills.  In contrast, \citet{COLOMBO2019} using the same ESCO version, but treat soft skills separate from hard skills. Most of the publications used all subcategories as skills without differentiating \citep{clavié2023large, gnehm-etal-2022-fine, decorte2023extreme}.

Beyond these, there are other skill bases, such as the Russian professional standard in \citet{Botov.2019} or the Chinese Occupation Classification Grand Dictionary used in \citet{Cao.2021, Cao2021b}. Additionally, non-official skill bases exist, like the list of 1K soft skills in \citep{Sayfullina.2018} or LinkedIn's in-house taxonomy for skill extraction~\citep{shi.2020}. In general, for transparency and reproducibility, it is helpful to state which subset of fine-grained labels \(L\) of the skill base (\(B\)) and which skill base version is used.

\paragraph{Leveraging Automated Tools}
Some studies leverage automated tools like AutoPhrase \citep{Shang.2018} or Microsoft Azure Analytics Service for NER for initial skill term detection, followed by manual verification and refinement \citep{Yao.2022, Kortum22}. Also \citet{vermeer.2022} extract parts of their training data using an automated tool, while others are taken from a skill base.\footnote{\url{https://www.textkernel.com/de/}} Lastly, \citet{Gugnani_Misra_2020} employ an IBM tool for skill identification, which forms a part of a larger skill identification framework.\footnote{\url{https://www.ibm.com/products/natural-language-understanding}} While  some previous work did not apply manual verification \citep{Gugnani_Misra_2020, vermeer.2022}, we recommend it to reduce automation bias from the tool impacting the data.

\paragraph{Definition through Labeling} 
Domain experts play a crucial role for labeling data and therefore impact how the definition of skills is put into work \citep{shi.2020, Tamburri.2020, beauchemin2022fijo}. \citet{Tamburri.2020} additionally provide a codebook with skill definitions to address ambiguities. \citet{shi.2020} used next skills identified by hiring experts and skills common among successful applicants as training data.
The study by \citet{bhola-etal-2020-retrieving} treat the companies filing the JPs as domain experts by using their labels (see also Section~\ref{BHOLA}). Besides domain experts, crowd workers and the people writing the guidelines for the workers oftentimes determine which terms are skills. Some studies do not mention who labels the data \citep{Wild.2021, Cao.2021, Botov.2019}. We suggest being clear about the labeling process and guidelines, making them public for transparency and re-use/standardization, and using domain experts if possible for accurate labeling.

\section{Data}

In this section, we provide a comprehensive description of publicly available datasets, with an overview in Table~\ref{tab:labeledData}.
\begin{table*}[th!]
\centering
\small
\begin{tabular}{llllllc}
\toprule
\textbf{Publication} & \textbf{Approach}  & \textbf{Granularity} & \textbf{Skill type} & \textbf{Use case} & \textbf{Size  }  & \faBook  \\
\midrule
{\citep{Sayfullina.2018}}& {Crowdsourced}  & {span-level} & {soft} & {\(I\)}  & {7411 spans} & {\faRemove} \\
{\citep{green-etal-2022-development}}  & {Crowdsourced} & {span-level} & {hard + soft} & {\(E_C\)} & {10,606 spans} &  {\faCheck}\\
{\citep{beauchemin2022fijo}}  & {Expert} & {span-level} & {soft}  & {\(E_C\)} & {47 JPs - 932 spans} & {\faRemove} \\
{\citep{zhang-etal-2022-skillspan}}& {Expert}  & {span-level} & {hard + soft}  & {\(E_C\)}  & {265 JP - 9,633 spans} & {\faCheck} \\
{\citep{zhang-etal-2022-kompetencer}}& {Expert} & {span-level} & {hard + soft} & {\(E_C\)+\(C_D\)}  & {60 JP - 920 spans} &  {\faCheck} \\
\citep{Decorte.2022}  & {Manual} & {span-level} & {hard + soft} & {\(I\)+\(C_D\)} & {1,618 spans} &  {\faCheck}\\
\citep{gnehm-etal-2022-evaluation} & {Expert}  &{span-level} &  {hard + soft}   & {\(E_C\)+\(C_D\)}  & {10,995 spans} &  {\faRemove} \\
{\citep{bhola-etal-2020-retrieving}} & {Skill Inventory}  & {document-level} & {unknown}  & {\(C_E\)}  & {20,298 JP}  & {\faRemove} \\

\bottomrule
\end{tabular}
\caption{Overview of publicly-available labeled datasets. \faBook{} indicates if the authors used guidelines (not necessarily publicly available).}
\label{tab:labeledData}
\end{table*}

\paragraph{SAYFULLINA} by \citet{Sayfullina.2018} \label{SAYFULLINA} is a dataset derived from a publicly available Kaggle dataset, containing JPs from within the UK and representing a variety of sectors.\footnote{\url{https://www.kaggle.com/datasets/airiddha/trainrev1/?select=Train_rev1.csv}} The authors retrieved soft skill spans by exact matching with a list of 1,072 soft skills. Each identified span is accompanied by up to 10 surrounding words. Crowdsourcing was used to determine whether the highlighted skill belongs to a job applicant. To ensure reliability, the workers were tested on a small set of JPs and each snippet was evaluated by at least three workers. This process led to a dataset with high class imbalance due to more positive examples. To counter this, additional skill spans were added, including those usually not describing candidates (marked as negative) and those consistently labeled positive.

\paragraph{GREEN} by \citet{green-etal-2022-development} uses the same Kaggle dataset as SAYFULLINA. The labeling was done via crowdsourcing, they did not use experts but only workers who passed a test were included, and encouraged to follow the guidelines. Apart from the ``Skill'' label capturing hard and soft skills, the labels ``Occupation'', ``Domain'', ``Experience'', ``Qualification'', and ``None'' are used in a BIO scheme. The authors reduced errors by label aggregation with a preference towards labels from higher-performing workers. Additionally, they reclassified specific ``Experience'' spans, as ``Skill'' spans, and manually split multi-term spans into separate spans. 

\paragraph{FIJO} by \citet{beauchemin2022fijo} was created in partnership with Canadian insurance companies, and consists of cleaned and de-identified French JPs published between 2009 and 2020. The dataset focus on soft skills and includes 867 JPs with 47 annotated JPs, selected and annotated by a domain expert. The annotated spans are unevenly distributed across four classes: ``Thoughts'', ``Results'', ``Relational'', and ``Personal''.

\paragraph{SKILLSPAN} by \citet{zhang-etal-2022-skillspan} consists of the anonymized raw data and annotations of skill and knowledge spans from three JP datasets, one of which cannot be made publicly available due to its license. The available datasets are: 
\begin{itemize}
    \itemsep0em
    \item \textbf{HOUSE}: A static in-house dataset with different types of JPs from 2012-2020 and 
    \item \textbf{TECH}: The StackOverflow JP platform, consisting mostly of technical jobs collected between June 2020 and September 2021.
\end{itemize}
The development of the publicly available annotation guidelines involved an iterative process, starting with a few JPs and progressing through several rounds of annotation and refinement by three domain experts.

\paragraph{KOMPETENCER} by \citet{zhang-etal-2022-kompetencer} consists of Danish JPs with annotated skill and knowledge spans, see Table~\ref{tab:spans} in the Appendix. The same skill definitions, guidelines, and metrics as in SKILLSPAN are used for annotation. This dataset can be used for skill extraction with coarse labels, but the authors have also added fine-grained annotations to evaluate a classification with the ESCO taxonomy. For fine-grained annotations, they query the ESCO API with the annotated spans and use Levenshtein distance to determine the relevance of each obtained label. Then, the quality of these distantly supervised labels is assessed through human evaluation. They also repeated this process for the English SKILLSPAN dataset but only manually checked a sample for calculating statistics. 

\paragraph{DECORTE} by \citet{Decorte.2022} \label{DECORTE} is a variant of the SKILLSPAN dataset with annotated ESCO labels. They used the identified skill without the skill and knowledge labels, but they can be recreated by matching the dataset with SKILLSPAN, see Table~\ref{tab:spans} in the Appendix. Unlike in KOMPETENCER they manually matched the skills with fitting ESCO labels (if they exist) to create a gold standard. 

\paragraph{GNEHM-ICT} by \citet{gnehm-etal-2022-evaluation} is a Swiss-German dataset where they annotated for Information and Communications Technology (ICT)-related entity recognition. These could be ICT tasks, technology stack, responsibilities, and so forth. The used dataset is a combination of two other Swiss datasets namely the Swiss Job Market Monitor and an online job ad dataset~\cite{gnehm2020text, buchmann2022swiss}. There are around 25,000 sentences in the dataset.

\paragraph{BHOLA} by \citet{bhola-etal-2020-retrieving} \label{BHOLA}  was obtained from a government website\footnote{\url{https://www.mycareersfuture.gov.sg/}.} in Singapore. The preprocessing steps for this English language dataset include converting text to lowercase and removing stop words and rarely used words. The companies filing the JPs added skill labels, which are mapped to the whole JP document. This makes the dataset suitable for performing multi-label classification by predicting a set of required skills for a given JP.

\section{Methods}
\begin{table*}[t]
\centering
\small
\begin{tabular}{lllll}
\toprule
\textbf{Paper} & \textbf{Model} & \textbf{Skill Type} & \textbf{Granularity} & \textbf{Use Case}\\
\midrule
\citep{Fang.2023} & {Custom pre-trained LM} & {soft + hard} & {word-level} & {\(E_C\)}\\
\citep{goyal-etal-2023-jobxmlc} & {FastText skip-gram, GNN} & {unknown} & {word-level} & {\(C_E\)}\\
\citep{clavié2023large} & {GPT-4} & {soft + hard} & {span-level} & {\(C_E\)} \\
\citep{li2023skillgpt} & {XMLC - LLM} & {soft + hard} & {document-level} & {\(C_E\)}\\ 
\citep{decorte2023extreme} & {GPT-3.5} & {soft + hard} & {sentence-level} & {\(C_E\)}\\ 
\citep{zhang-etal-2023-escoxlm} & {Multilingual XLM-R} & {soft + hard} & {span-level} & {\(E_C\)}\\
\citep{Decorte.2022} & {RoBERTa} & {soft + hard} & {sentence-level} & {\(C_E\)}\\
\citep{zhang2022skill} & {RoBERTa, JobBERT} & {soft + hard} & {span-level} & {\(C_D\)} \\
\citep{gnehm-etal-2022-fine}  & {JobBERT-de, SBERT} & {soft + hard} & {span-level}& {\(E_C\) + \(C_D\)}\\
\citep{zhang-etal-2022-kompetencer} & {BERTbase , DaBERT} & {soft + hard} & {span-level} & {\(C_E\)} \\
\citep{beauchemin2022fijo} & {Bi-LSTM, CamemBERT} & {soft} & {span-level} & {\(E_C\)}\\
\citep{Yao.2022} & {BERT, word2vec} & {unknown} & {word-level} & {\(I\)}\\
\citep{anand2022required} & {LaBSE model} & {soft + hard} & {title} & {\(C_E\)} \\
\citep{vermeer.2022} & {RobBERT} & {soft + hard} &  {document-level} & {\(C_E\)} \\
\citep{Wild.2021} & {BERT, spaCy} & {soft + hard} & {span-level} & {\(I\)} \\
\citep{Khaouja.2021b} & {Sent2vec, SBERT} & {soft + hard} & {sentence-level} & {\(C_E\)}\\
\citep{Cao2021b} & {BERT-BiLSTM-CRF} & {soft + hard} & {span-level} & {\(I\)} \\
\citep{Cao.2021} & {BERT-BiLSTM-CRF} & {soft + hard} & {span-level} & {\(I\)}\\
\citep{Li.2020} & {Deep Averaging Network,  FastText} & {unknown}& {span-level} & {\(C_E\)} \\ 
\citep{Tamburri.2020} & {BERT Multilingual Cased} & { soft + hard} & {sentence-level} & {\(I\)} \\
\citep{bhola-etal-2020-retrieving} & {BERTbase} & {unknown} & {document-level} & {\(C_E\)} \\
\citep{Gugnani_Misra_2020} & {Word2vec} & {soft + hard }& {span-level }  & {\(I\)} \\
\citep{Botov.2019} & {Word2vec} & {unknown } & {span-level} & {\(C_E\)} \\
\citep{Jia.2018} & {LSTM} & {unknown} & {word-level} & {\(I\)}\\
\citep{Sayfullina.2018} & {CNN, LSTM, HAN} & {soft} & {span-level} & {\(I\)} \\
\citep{Javed_Hoang_Mahoney_McNair_2017} & {Word2vec} & {soft + hard} & {span-level} & {\(C_E\)}\\
\bottomrule
\end{tabular}
\caption{Publications regarding neural skill extraction and classification. The skill type was not always explicitly mentioned in some cases it's derived from examples given in the paper.}
\label{tab:accents}
\end{table*}

In this section, we survey methods for skill extraction and classification. As in Section~\ref{chap:Terminology} the goal of the extraction is to identify skill spans with (\(E_C\)) or without coarse labels (\(I\)). The classification section covers direct classification methods (\(C_D\)) and classification methods with extraction (\(C_E\)), both aim to retrieve fine-grained skill labels.


\subsection{Skill Extraction}\label{chap:Extraction}
This chapter delineates the evolution of skill extraction methodologies, grouped into three categories: skill identification as span labeling, skill identification through binary classification, and skill extraction with coarse span labels. Starting with LSTM neural networks in 2018 the methods in all three sub-chapters used after the introduction of BERT \citep{devlin-etal-2019-bert} in 2019 heavily BERT and BERT-based models. Recent advancements continue to diversify the landscape, integrating a broader array of language models (LMs).


\subsubsection{Skill Identification as Span Labeling}
In this category approach skill identification as a span labeling task. The primary objective is to accurately identify skill spans, encompassing both the identification of the relevant skill phrases and their precise boundaries.
\citet{Jia.2018} are the first to use sequence tagging for identifying skills from JPs in 2018. The authors use a pre-trained LSTM neural network \citep{lample-etal-2016-neural} for identifying skill terms on the word-level. \citet{Tamburri.2020} also employed binary classification, but at the sentence-level, using a Dutch JP dataset. Their best-performing model, BERT Multilingual Cased, was fine-tuned on expert-annotated JP sentences, suggesting potential improvement with more data and optimization. Further publications retrieve embeddings using a pre-trained BERT model \citep{Wild.2021, Cao.2021, Cao2021b}. Notably, \citet{Cao2021b} and \citet{Cao.2021} combine BERT's pre-trained vectors with a Bi-LSTM and a CRF layer for finer entity classification. This approach aligns with previous research demonstrating the efficacy of a CRF layer in NER tasks \citep{souza2020portuguese}. In~\citet{zhang-etal-2023-escoxlm}, they further built upon the domain-adaptive pre-training paradigm~\citep{gururangan-etal-2020-dont}. They make use of the ESCO taxonomy~\citep{Vrang.2014} and integrate this in a multilingual XLM-R model~\cite{conneau-etal-2020-unsupervised}, using this taxonomy-driven pre-training method, they introduce a new state-of-the-art for all skill identification benchmarks. For analysis, they show that performance increases especially for skills that are shorter in length, due to ESCO skills also being shorter. 

In contrast to these single-model approaches, \citet{Gugnani_Misra_2020} adopted a multi-faceted methodology to predict the relevance of identified skill spans. Their methodology encompassed four modules: using part-of-speech (PoS) tagging, parsing sentences with skill bases (O*NET, Hope, and Wikipedia), leveraging a ready-made sequence tagging solution, and employing a pre-trained word2vec model for final score determination through cosine similarity.\footnote{\url{https://www.ibm.com/products/natural-language-understanding}.}

\subsubsection{Skill Identification as binary Classification Task}
In this category, skill identification is framed as a binary classification task. The focus is on determining whether a given sequence either constitutes or contains a (specific) skill. 
The task in \citet{Sayfullina.2018} differs from the other publications. They extract skill spans by exact match and aim to decide whether skill spans refer to a candidate or something else, like a company. They experiment with various classifiers and input representations, such as Soft Skill Masking, Embedding, and Tagging, finding the LSTM classifier with skill tagging most effective on their dataset. \citet{Tamburri.2020} employed binary classification at the sentence-level to determine if it contains a skill.  Their best-performing model, BERT Multilingual Cased, was fine-tuned on expert-annotated JP sentences using a Dutch JP dataset. \citet{Yao.2022} classify individual words as skill-related or not. They split JPs into individual words, analyzing each through character-level and word-level encoders, integrating linguistic features like POS tags and capitalization. Their initial training employs AutoPhrase \citep{Shang.2018} for automatic skill term identification, followed by manual verification and expert-labeled samples. The model is further refined using Positive-Unlabeled learning, where the classifier's predictions on unlabeled data help expand the skill base for continuous adaptation.


\subsubsection{Skill Extraction with Coarse Labels}

This section explores advancements in skill extraction with coarse labels, where each publication extract spans from two to four different categories. The studies of \citet{gnehm-etal-2022-fine} and \citet{zhang-etal-2022-skillspan} both utilize sequence tagging-based models. \citet{gnehm-etal-2022-fine} focusing on iterative training and annotation with jobBERT-de, a German LM tailored for JPs. \citet{zhang-etal-2022-skillspan} compare BERT-based \citep{devlin-etal-2019-bert} and SpanBERT-based \citep{joshi-etal-2020-spanbert} models, highlighting the importance of domain adaptation. On the other hand, \citet{beauchemin2022fijo} and \citet{Fang.2023} delve into the intricacies of training and optimizing LMs for skill extraction. \citet{beauchemin2022fijo} examine the sensitivity of Bi-LSTM and CamemBERT~\cite{martin-etal-2020-camembert} models to training data volume, with CamemBERT unfrozen yielding the highest mean token-wise accuracy. \citet{Fang.2023} introduce RecruitPro, a specialized model for skill extraction from recruitment texts, employing innovative techniques for dealing with data noise and label imbalances. Collectively, these papers emphasize the need for tailored approaches and continuous innovation in model development.

\subsection{Skill Classification}\label{chap:Classification}
While skill standardization can be achieved through classification, other methods such as clustering \citep{moreno.2019, Lukauskas.2023}, matching n-grams based on string similarity \citep{BOSELLI.2018}, or identifying semantically similar skills \citep{moreno.2019, COLOMBO2019, Grueger.19} also lead to standardized skill spans. These methods simplify the variety and quantity of skill spans without assigning standardized labels. Transitioning from these methods, we now focus on skill classification, a crucial step for assigning standardized labels to effectively organize and understand skills. Most publications skip a traditional extraction and match the JPs directly to the skill base (\(C_E\)), which can be seen as skill extraction against a skill base. Exceptions are \citet{gnehm-etal-2022-fine}, which perform extraction of skill spans with coarse labels before the fine-grained classification step, and \citet{zhang-etal-2022-kompetencer} who rely on prior work for extraction and focus solely on the matching of skill spans to ESCO (\(C_D\)). We divide the publications by methodology into those that match based on semantic similarity and those using extreme multi-label classification to solve the matching task.

\subsubsection{Similarity-based Approaches}
The publications with similarity-based approaches split the JPs into sentences or n-grams before matching them. All of the following publications use skill embedding methods, which can be seen as an advancement of the skill count methods (Section~\ref{ch:Introduction}). The advances in text embeddings over time are reflected in the scope of the approaches. While  \citet{Javed_Hoang_Mahoney_McNair_2017} and \citet{Botov.2019} improve the matching using word2vec embeddings\citep{mikolov2013efficient}, later \citet{Li.2020} use FastText \citep{bojanowski2017enriching} leveraging sub-word information to handle  out-of-vocabulary words and capture more detailed semantic and syntactic information. \citet{Khaouja.2021b} compare using sent2vec trained on Wikipedia sentences, and SBERT \citep{reimers-gurevych-2019-sentence} trained on millions of paraphrase sentences for embeddings. Moreover, \citet{zhang2022skill} uses LMs like RoBERTa and JobBERT to match n-grams from JP sentences with the ESCO taxonomy. They also experiment with context and frequency-aware embeddings. 
\citet{gnehm-etal-2022-fine} performed direct skill extraction using context-aware embeddings and the SBERT model similar to \citet{zhang2022skill}, additionally they contextualize skill areas within spans and ontology terms using their hierarchical structure. The study explores techniques to enhance BERT model similarity, including in-domain pretraining, transformer-based sequential denoising auto-encoder (TSDAE; \citealp{wang-etal-2021-tsdae-using}) for domain-specific terminology, and Siamese BERT Networks for training sentence embeddings~\citep{reimers-gurevych-2019-sentence}. They further leverage MNR loss in Siamese networks \citep{henderson2017efficient}, using ontology data to create positive text pairings for better label matching. SkillGPT~\cite{li2023skillgpt} is the first tool to use an LLM for the matching task, they convert ESCO entries into structured documents, which are vectorized by the LM. Then, they summarize the input text, and use an embedding of the summary to retrieve the closest ESCO entries.

\subsubsection{Extreme Multi-label Classification Approaches}

\citet{bhola-etal-2020-retrieving} were the first to formulate skill extraction against a skill base as an extreme multi-label classification (XMLC). They classify multiple skill labels per document using the labels of the BHOLA dataset (around 2500 labels) as a skill base. Their BERT–XMLC framework, involves a Text Encoder that uses the pre-trained BERTbase model to convert JP texts into dense vector representations, a Bottleneck Layer that reduces overfitting by compressing these representations~\citep{Liu.2017} and subsequently a fully connected layer for multi-label classification of the skills. Enhancements include focusing on semantic skill label representation and skill co-occurrence, using bootstrapping to augment training data, and improve skill correlation capture. Their model outperformed XMLC baselines. \citet{vermeer.2022} adapted this approach for using RobBERT and additional linear layers, validating on BHOLA and a non-public Dutch dataset. Similarly, \citet{anand2022required} extended the model to predict skill importance using LaBSE-encoded~\cite{feng-etal-2022-language} job titles, ranking skills from an in-house database based on a 0-1 scale of importance.

Subsequent publications have concentrated on XMLC for skill extraction and classification using the ESCO taxonomy with around 13000 labels. For a pure skill classification for already identified skill spans \citet{zhang-etal-2022-kompetencer} use distant supervision by querying the ESCO API for the fine-grained skill labels. For model training, they employ zero-shot cross-lingual transfer learning techniques using various BERT models and fine-tune them on Danish JPs. The effectiveness of the models is tested on an adapted version of SKILLSPAN and KOMPETENCER. The same year \citet{Decorte.2022} addressed the XMLC task on the sentence-level, again using distant supervision with the ESCO taxonomy.  They enhance binary skill classifier training with three negative sampling strategies, involving siblings in ESCO hierarchy, Levenshtein distance, and cosine similarity of RoBERTa-encoded skill names. Their model employs a frozen pre-trained RoBERTa with mean pooling for sentence representation, followed by separate binary classifiers for each skill, evaluated on DECORTE.

As for the similarity-based approaches, LLMs are prominent in recent XMLC approaches. Unlike \citet{li2023skillgpt}, \citet{decorte2023extreme} use the LLM solely during training to reduce latency and enhance reproducibility. They create a synthetic training dataset using the LLM, then optimize a bi-encoder through contrastive training, to effectively represent both skill names and corresponding sentences in close proximity within the same space. This method outperforms the distance supervision baseline by \citet{Decorte.2022} (see Table~\ref{tab:scores}). Similarly, \citet{clavié2023large} treat the skill extraction and classification task as individual binary classification problems, using GPT-3.5 like \citet{decorte2023extreme} but generating more spans per skill for synthetic training. They propose two extraction methods: one using linear classifiers for each skill, employing hard negative sampling \citep{robinson2021contrastive} for improved skill differentiation, and another based on similarity, utilizing E5-LARGE-V2 embeddings \citep{wang2022text} for cosine similarity calculations between JP extracts and ESCO labels or synthetic sentences. Potential skills are then reranked using an LLM. In evaluations using the DECORTE dataset, their methods achieved high performance with GPT-4, though results with GPT-3.5 were lower than \citet{decorte2023extreme}, see Table~\ref{tab:scores} in the Appendix.

\citet{goyal-etal-2023-jobxmlc} present JobXMLC, a unique framework for the XMLC task, distinct from the prevailing methods. JobXMLC integrates a job-skill graph to represent job-skill interconnections, utilizes a GNN for multi-hop embeddings from the graph's structure, and incorporates an extreme classification system with skill attention based on skill frequency in the dataset. The framework's effectiveness is validated on the BHOLA and a proprietary StackOverflow dataset, see Table~\ref{tab:scores} in the Appendix.

\section{Conclusions and Future Directions}
Recent publications indicate two emerging trends in skill extraction. Firstly, extracting skills against skill bases like ESCO is gaining popularity, facilitating cross-industry and regional comparisons. Secondly, LLMs are increasingly applied in skill extraction and classification, proving particularly advantageous due to the scarcity of training data in this domain.

Future research in skill extraction and classification could focus on emerging skills and the extraction of implicit skills. Methods like those by \citet{Javed_Hoang_Mahoney_McNair_2017} and \citet{Khaouja.2021b} update skill bases with emerging technologies and frequently used keywords, but evaluating these remains difficult without a standard benchmark. The challenge of extracting implicit skills, not directly stated in job postings, is also gaining attention. Techniques include prompting LLMs to generate training data with implied skills \citep{clavié2023large} and using complete sentences to encompass both explicit and implicit skills \citep{Decorte.2022, decorte2023extreme}. However, these methods need thorough evaluation, presenting an open field for future exploration.

\section*{Limitations}
A limitation that should be considered is that only publications in the English language (although data was from multiple languages) were surveyed in this paper. Second, to allow for a deeper focus publications regarding topic modeling were excluded even if they used deep-learning-based methods.

\section*{Acknowledgements}
We thank the reviewers for their insightful feedback. ES acknowledges financial support with funds provided by the German Federal Ministry for Economic Affairs and Climate Action due to an enactment of the German Bundestag under grant 46SKD127X (GENESIS). MZ is supported by the Independent Research Fund Denmark (DFF) grant 9131-00019B and BP is supported by ERC Consolidator Grant DIALECT 101043235.

\bibliography{custom}
\clearpage
\appendix

\section{Appendix}
\FloatBarrier 
\subsection{Terminology Example} \label{term_example}
In Table~\ref{tab:terminology}, we present an example sentence for better terminology understanding.

\begin{table}[h]
\setlength{\tabcolsep}{3pt}
    \begin{tabular}{lllllllllll}
 & Familiar& with & building & tests & in & python \vspace{.2cm} \\ 
$I$: & O & O & B & I& O & B\\ 
$E_C$: & O& O& B$_{skill}$ & I$_{skill}$ & O & B$_{knowl.}$\\
$C_D / C_E$: & \multicolumn{6}{l}{``Python (computer programming)'', ``
plan ''} \\
& \multicolumn{6}{l}{``software testing''}\\
    \end{tabular}
    
    \caption{An example with annotations for the different tasks described in Section~\ref{chap:Terminology}. For skill classification ($C$), we used the ESCO taxonomy in this example, and for skill extraction with coarse labels ($E_C$) we follow the guidelines of SkillSpan~\cite{zhang-etal-2022-skillspan}}
    \label{tab:terminology}
\end{table}

\begin{table*}[t]
\centering
\begin{tabular}{lcc}
\hline
\textbf{Source} & {\textbf{\# Skill Spans}} & \textbf{{\# Knowledge Spans}} \\
\hline
SKILLSPAN - HOUSE & 2,146 & 1,418\\
DECORTE - HOUSE & 509* & 210* \\
SKILLSPAN - TECH& 2,241 & 3,828\\
DECORTE - TECH & 419 & 480*  \\
KOMPETENCER & 665 & 255 \\
\hline
\end{tabular}
\caption{Number of labeled spans. The star * indicates, that two values found in the Decorte HOUSE test dataset (tagged as knowledge) were actually from the Skillspan TECH dataset; eight values found in the Decorte TECH test dataset (four skill spans, four knowledge spans) were actually from the Skillspan HOUSE dataset. \\}
\label{tab:spans}
\end{table*}

\subsection{Number of Skill and Knowledge Spans}
In Table~\ref{tab:spans}, we show the number of labeled spans for skills and knowledge in the SKILLSPAN~\cite{zhang-etal-2022-skillspan}, DECORTE~\cite{Decorte.2022}, and KOMPETENCER~\cite{zhang-etal-2022-kompetencer} dataset.

\subsection{Scores of Selected Models}
In Table~\ref{tab:scores}, we display the scores of recent LMM-based approaches on the DECORTE~\cite{Decorte.2022} dataset for comparison. Furthermore, we show results of \citet{zhang-etal-2023-escoxlm, goyal-etal-2023-jobxmlc} and \citep{bhola-etal-2020-retrieving} on the BHOLA~\cite{bhola-etal-2020-retrieving} dataset. 

\begin{table*}[t!]
\centering
\resizebox{\textwidth}{!}{
\begin{tabular}{lcccccccccc}
\hline
\textbf{Model} & \textbf{Source} & \multicolumn{3}{c}{\textbf{HOUSE*}} & \multicolumn{3}{c}{\textbf{TECH*}} & \multicolumn{3}{c}{\textbf{BHOLA}} \\
\cline{3-11} 
 &  & \textbf{MRR} & \textbf{RP@5} &
 \textbf{RP@10} &
 \textbf{MRR} & \textbf{RP@5}& \textbf{RP@10} &
 \textbf{MRR} & \textbf{R@5}& \textbf{R@10} \\
\hline
{\(Classifier^{neg} \)} & \citep{Decorte.2022}  & 0.299 & 30.82 & 38.69 & 0.326 & 31.71 & 39.09 & N/A & N/A & N/A \\
{\(GPT sentences^{aug} \)} & \citep{decorte2023extreme} & 0.428 & 45.74 & N/A  & 0.529& 54.62 & N/A & N/A & N/A & N/A \\
\(GPT3.5 Re-ranking \) & \citep{clavié2023large} & 0.427 & 43.57 & 51.44 & 0.488 & 52.50 & 59.75 & N/A & N/A & N/A \\
\(GPT4 Re-ranking \) & \citep{clavié2023large} & 0.495 & 53.34 & 61.02 & 0.537 & 61.50 & 68.94 & N/A & N/A & N/A \\
\(BERT–XMLC +CAB \) & \citep{bhola-etal-2020-retrieving} & N/A & N/A & N/A & N/A & N/A & N/A & 0.9049 & 21.67 & 40.49 \\
{\(JobXMLC \)} & \citep{goyal-etal-2023-jobxmlc} & N/A & N/A & N/A & N/A & N/A & N/A & 0.90 & 18.29 & 32.33 \\
\(ESCOXML-R \) & \citep{zhang-etal-2023-escoxlm} & N/A & N/A & N/A & N/A & N/A & N/A & 0.907 & N/A & N/A \\
\hline
\end{tabular}
}
\caption{Scores of selected models on the benchmarking datasets DECORTE and BHOLA.}
\label{tab:scores}
\end{table*}

\end{document}